\pdfoutput=1
\documentclass[letterpaper]{article} 
\usepackage{aaai23}  
\usepackage{times}  
\usepackage{helvet}  
\usepackage{courier}  
\usepackage[hyphens]{url}  
\usepackage{graphicx} 
\usepackage{natbib}  
\usepackage{caption} 
\usepackage{algorithm}
\usepackage{algorithmic}

\usepackage{newfloat}
\usepackage{listings}
\DeclareCaptionStyle{ruled}{labelfont=normalfont,labelsep=colon,strut=off} 
\floatstyle{ruled}
\newfloat{listing}{tb}{lst}{}
\floatname{listing}{Listing}
\usepackage{multirow}
\usepackage{booktabs}

\title{TA-DA: Topic-Aware Domain Adaptation for \\ Scientific Keyphrase Identification and Classification (Student Abstract)}
\author{
Răzvan-Alexandru Smădu\textsuperscript{\rm 1}, George-Eduard Zaharia\textsuperscript{\rm 1}, Andrei-Marius Avram\textsuperscript{\rm 1}, \\ Dumitru-Clementin Cercel\textsuperscript{\rm 1}, Mihai Dascalu\textsuperscript{\rm 1}, Florin Pop\textsuperscript{\rm 1,\rm 2}
}
\affiliations{
    \textsuperscript{\rm 1}Faculty of Automatic Control and Computers, University Politehnica of Bucharest \\
    \textsuperscript{\rm 2}National Institute for Research and Development in Informatics - ICI Bucharest, Romania \\
    \{razvan.smadu,george.zaharia0806,andrei\_marius.avram\}@stud.acs.upb.ro\\
    \{dumitru.cercel,mihai.dascalu,florin.pop\}@upb.ro
}

\begin{document}

\maketitle

\begin{abstract}
Keyphrase identification and classification is a Natural Language Processing and Information Retrieval task that involves extracting relevant groups of words from a given text related to the main topic. In this work, we focus on extracting keyphrases from scientific documents. We introduce TA-DA, a Topic-Aware Domain Adaptation framework for keyphrase extraction that integrates Multi-Task Learning with Adversarial Training and Domain Adaptation. Our approach improves performance over baseline models by up to 5\% in the exact match of the F1-score.
\end{abstract}

\section{Introduction}

Scientific Keyphrase Identification and Classification (SKIC) refers to the labeling of relevant words from a given input scientific document, which the SemEval-2017 workshop~\citep{augenstein-etal-2017-semeval} introduced in Task 10. This task is related to Named Entity Recognition (NER), where the aim is to extract and classify the named entities from a given set of documents. SKIC can be more challenging than NER problems due to the lack of available annotated scientific publications~\citep{augenstein-etal-2017-semeval}.

\begin{figure}[!t]
\centering
\includegraphics[width=0.99\columnwidth]{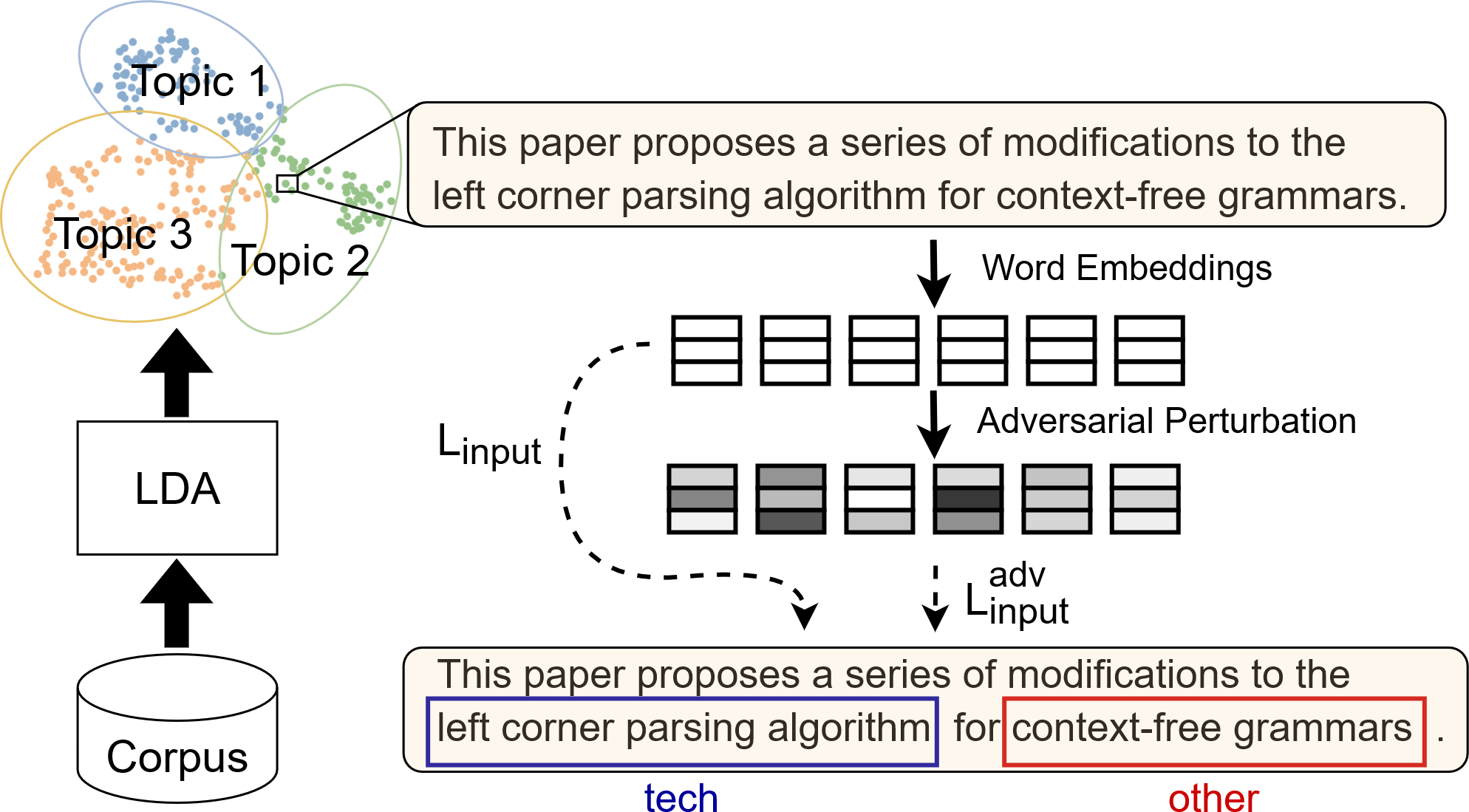}
  \caption{The training procedure for TA-DA employing domain adaptation and adversarial training.}
  \label{fig:Keyphrase_img.png}
\end{figure}

This work focuses on using domain adaptation through adversarial training and adversarial examples to address SKIC. We introduce a Topic-Aware Domain Adaptive (TA-DA) deep neural network framework that incorporates multi-task learning and adversarial learning. Figure~\ref{fig:Keyphrase_img.png} briefly presents the training procedure. Experiments showed that each component improves the performance of our model on all considered datasets. We summarize our main contributions as follows:

\begin{itemize}
    \item We propose a novel neural network architecture integrating multi-task learning, adversarial training, domain adaptation, and topic modeling for jointly addressing SKIC tasks;
    \item Different techniques for latent representations required for topic modeling (i.e., LSA, NMF, K-Means, LDA) are considered, and LDA provided the most promising overall results; 
    \item We show that domain adaptation improves performance compared to baselines, while adding adversarial examples further increases the F1-scores.
\end{itemize}

\section{Methodology}

\subsubsection{Data Representation.}
We cast the problem of SKIC to a tagging task.
Given a document $d = \{w_1, ...,w_n\}$ with $n$ words, our goal is to identify and classify keyphrases by outputting a sequence $\{y_1, ..., y_n\}$, where each $y_i$ is a class from the BIO schema \citep{lample-etal-2016-neural}. 

\subsubsection{LDA for Topic Modeling.}
Our intuition is to extract latent topics related to the domains and combine them with the labels of examples. Latent Dirichlet Allocation (LDA) \citep{10.5555/944919.944937} is a frequently employed unsupervised method to determine latent topics from a corpus. It associates each document with a mixture of topics as probability distributions of words that are thematically related based on tight co-occurrence. 

\subsubsection{Neural Network Architecture.}
We propose a neural network architecture starting with a pre-trained SciBERT \cite{beltagy-etal-2019-scibert} that generates contextualized embeddings. To further enhance the latent space, this representation is fed into a stack of BiLSTM layers, whose outputs are concatenated along with the SciBERT's output. We use multi-task learning to incorporate branches for each task that share the feature representation. After the feature extractor, fully connected layers are used for keyphrase identification (KI) and classification (KC) tasks. For KI, the output is a linear-chain CRF, which models the probability of having the output label given the latent representation. For KC, a softmax activation function is considered to model a probability distribution across classes. We incorporate adversarial training by including a domain discriminator linked via a gradient reversal layer \citep{ganin2015unsupervised} to the feature extractor. The output is computed using a softmax layer that models a probability distribution across domains, i.e. the topics extracted using LDA.

\subsubsection{Adversarial Learning.}
\label{section:adv_training}

We perform domain adaptation through adversarial training~\citep{ganin2015unsupervised}. In this configuration, adversarial training seeks the saddle point that minimizes the expected likelihood of the loss associated with the label predictor while maximizing the likelihood for the discriminator.
Additionally, we employ adversarial examples via Fast Gradient Sign Method~\citep{goodfellow2015explaining} that adds a small value to the input  (in our case, the latent space generated by SciBERT) proportional to the gradient in order to create more robust models. The new input is defined as follows:

\begin{equation}
\label{eq:fgsm}
x_{adv} = x + \epsilon sign(\nabla_{x} L(\theta, x, y))
\end{equation}
where $\epsilon$ is the perturbation factor.

\subsubsection{Optimization.}
The optimization process involves minimizing the negative log-likelihood of the CRF output and the categorical cross-entropy of the softmax outputs. The total loss is described by:
\begin{equation}
\label{eq:total_loss}
L_{total} = L_{tag} + L_{class} + \lambda L_{da}
\end{equation}
where $L_{tag}$ is the KI loss, $L_{class}$ is the KC loss, $\lambda$ controls the degree of enforced domain adaptation, and $L_{da}$ is the domain adaptation loss.
In the case of using adversarial examples, the loss $ L_{total}^{adv} $ is similar to (\ref{eq:total_loss}), and both losses for each pass are added during the training step.

\section{Results}

\subsubsection{Datasets.}
We evaluate our approach on three datasets: SemEval-2017 Task 10 Dataset~\cite{augenstein-etal-2017-semeval}, SciERC (SciIE) Dataset~\cite{luan2018multitask}, and ACL RD-TEC 2.0 Dataset~\cite{qasemizadeh-schumann-2016-acl}.

\subsubsection{Experimental Results.}
The results are shown in Table \ref{tab:ablation}. 
Our approach (i.e., DA-Adv) obtains the best or second-best scores concerning the baselines. Although domain adaptation may hinder performance by 1 to 2\%, we observe that adversarial examples enhance the model's performance in every scenario.
Additionally, including LSA instead of LDA for topic modeling (i.e., DA-LSA), we observe similar performances, albeit on average, LDA performs better. Conversely, NMF (i.e., DA-NMF) and K-Means (i.e., DA-KM) hinder performances when employed as topic models.
Compared with LSTM-CRF~\cite{lample-etal-2016-neural}, our approach is comparable or performs better.
Also, our model has considerably fewer trainable parameters (2.4M) than SciBERT~\cite{beltagy-etal-2019-scibert} (109M) and improved results when compared with similar approaches (i.e., MTL-LSTM~\cite{augenstein2017multi}).

\begin{table}[hbt]
    \centering
    \small
    \begin{tabular}{p{1.5cm}cccccc}
      \toprule
      \multicolumn{1}{c}{\multirow{2}{*}{\textbf{Model}}} & \multicolumn{2}{c}{\textbf{SemEval}}                                       & \multicolumn{2}{c}{\textbf{SciIE}}                                         & \multicolumn{2}{l}{\textbf{ACL RD-TEC}}                                    \\
      \multicolumn{1}{c}{}                                & \multicolumn{1}{l}{\textbf{KI}} & \multicolumn{1}{l}{\textbf{KIC}} & \multicolumn{1}{l}{\textbf{KI}} & \multicolumn{1}{l}{\textbf{KIC}} & \multicolumn{1}{l}{\textbf{KI}} & \multicolumn{1}{l}{\textbf{KIC}} \\ \toprule
      DA-Adv                         & 53.6             & \underline{42.6}      & 78.4             & 65.9             & \underline{82.5} & \textbf{70.0}    \\
      DA-LSA                     & 53.2             & 41.5             & \underline{78.9}     & \underline{66.8} & \textbf{82.7}    & \underline{69.5} \\
      DA-NMF                     & 54.0             & 41.1             & 76.6             & 63.6             & 81.6             & 69.3             \\
      DA-KM                  & 49.7             & 38.0             & 77.7             & 65.3             & 81.5             & 69.0             \\ \midrule \midrule
      LSTM-CRF    & 58.6             & 33.7             & -                & -                & 74.4             & 35.5             \\
      MTL-LSTM        & \textbf{67.7}    & 38.0             & 72.3             & 58.1             & 81.9             & 58.5             \\
      SciBERT  & \underline{66.7} & \textbf{48.2}    & \textbf{81.0}    & \textbf{67.4}    & 79.8             & 65.1 \\
      \bottomrule
    \end{tabular}
    \caption{\label{tab:ablation} F1-scores on the test sets (the highest scores are bolded, and the second-highest scores are underlined).}
\end{table}
  
\section{Conclusions}

We proposed TA-DA, a topic-aware domain adaptation approach for tackling SKIC. We conducted multiple experiments and analyzed the performance impact of adversarial learning and topic modeling. We observed that the adversarially trained models outperform other models on one dataset, while considerable improvements are achieved in contrast to LSTM-CRF-based models. Moreover, we showed that domain adaptation and adversarial examples improved results over baselines.
In future work, we aim to study a hierarchical representation of the input to increase robustness further.

\section*{Acknowledgements}

The research has been funded by the University Politehnica of Bucharest through the PubArt program.

\small
\bibliography{aaai23}

\end{document}